\documentclass[twoside,11pt]{article}
\usepackage{jair, theapa, rawfonts}

\jairheading{74}{2022}{1201-1223}{02/2022}{07/2022}
\ShortHeadings{CoLLIE: Continual Learning of Language Grounding}
{Skantze \& Willemsen}
\firstpageno{1201}

\usepackage{amsmath,amsfonts,bm}

\def\eqref#1{equation~\ref{#1}}

\def\1{\bm{1}}

\DeclareMathAlphabet{\mathsfit}{\encodingdefault}{\sfdefault}{m}{sl}
\SetMathAlphabet{\mathsfit}{bold}{\encodingdefault}{\sfdefault}{bx}{n}

\def\sR{{\mathbb{R}}}

\usepackage{url}

\usepackage{graphicx}
\usepackage{wrapfig}
\graphicspath{ {images/} }

\begin{document}

\title{CoLLIE: Continual Learning of Language Grounding from Language-Image Embeddings}

\author{Gabriel Skantze and Bram Willemsen  \\
Department of Speech Music and Hearing\\
KTH Royal Institute of Technology\\
Stockholm, Sweden \\
\texttt{\{skantze,bramw\}@kth.se} 
}

\author{\name Gabriel Skantze \email skantze@kth.se \\
       \name Bram Willemsen \email bramw@kth.se \\
       \addr KTH Royal Institute of Technology,\\
       Stockholm, Sweden}


\maketitle

\begin{abstract}
This paper presents CoLLIE: a simple, yet effective model for continual learning of how language is grounded in vision. Given a pre-trained multimodal embedding model, where language and images are projected in the same semantic space (in this case CLIP by OpenAI), CoLLIE learns a transformation function that adjusts the language embeddings when needed to accommodate new language use. This is done by predicting the difference vector that needs to be applied, as well as a scaling factor for this vector, so that the adjustment is only applied when needed. Unlike traditional few-shot learning, the model does not just learn new classes and labels, but can also generalize to similar language use and leverage semantic compositionality. We verify the model's performance on two different tasks of identifying the targets of referring expressions, where it has to learn new language use. The results show that the model can efficiently learn and generalize from only a few examples, with little interference with the model's original zero-shot performance. 
\end{abstract}

\section{Introduction}
\label{section-introduction}
Any artificial agent interacting with an environment, using vision, and communicating with other agents (such as humans), using language, needs to be able to ground the meaning of language with the visual properties of the environment. One approach to this problem is to project vision and language into a joint semantic embedding space \shortcite{Frome2013,Bruni2014}. In such a model, a visual stimulus and a language construct that have similar representations are supposed to have similar meanings. In order to name a given object with certain visual features, the agent should try to generate a referring expression that has a similar embedding as the visual features of the object, and in order to understand what a referring expression is denoting, it should look for objects that have a similar visual feature embedding as that of the referring expression. 

\begin{figure}[t]
\begin{center}
\includegraphics[width=0.9\textwidth]{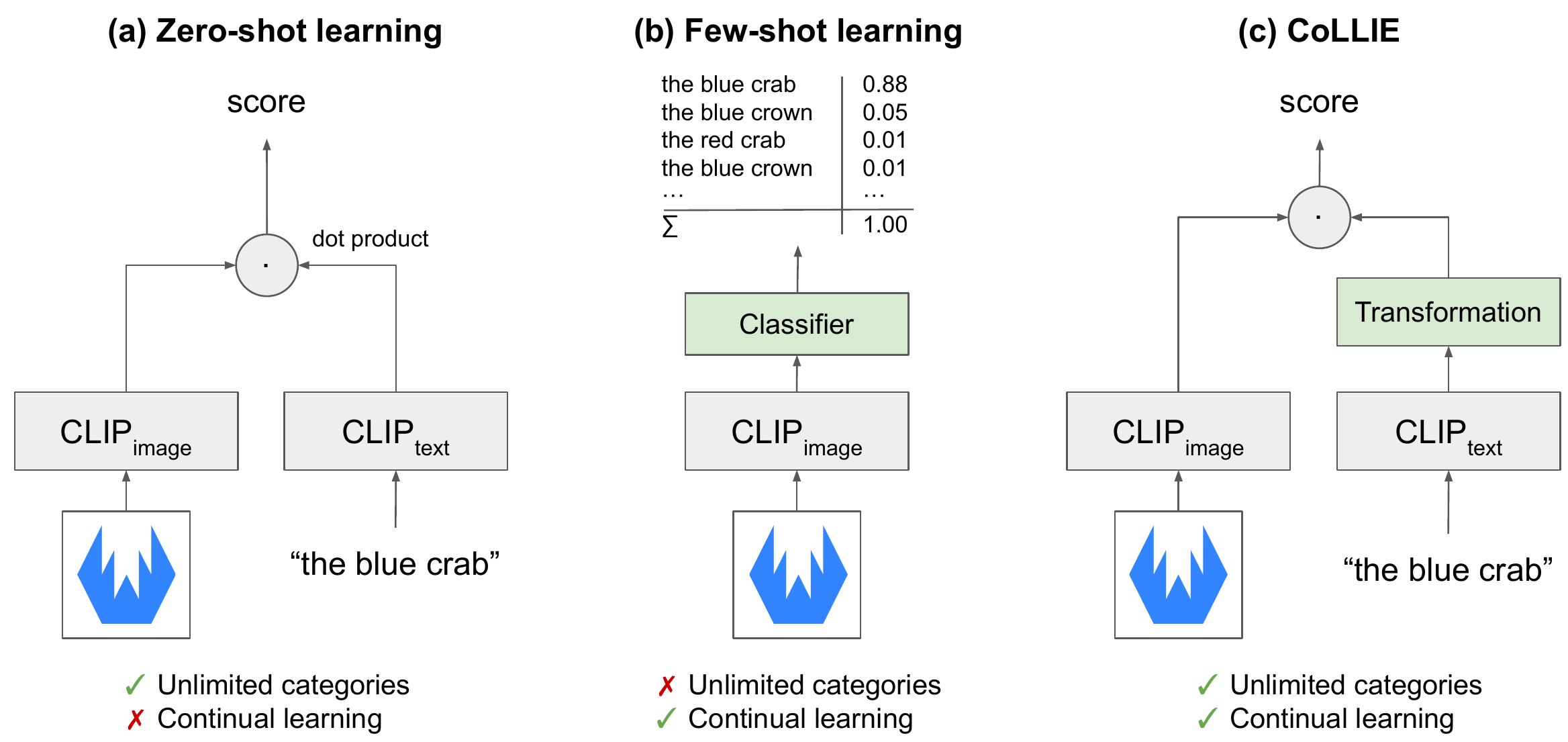}
\end{center}
\caption{Comparison of CoLLIE to Zero-shot and Few-shot learning. Green boxes show where continual learning is taking place.}
\label{fig:comparison}
\end{figure}

Recent developments in multimodal representation learning using large amounts of data have given impressive results. An example of a model integrating language and vision is CLIP by OpenAI \shortcite{Radford2021}, which was trained using constrastive learning on 400 million pairs of images and their captions. Images and texts are embedded (separately) using state-of-the-art computer vision and language processing pipelines into a 512-dimensional vector. By calculating the dot product of the two embeddings, it is possible to determine how similar an image is to a text (or an image to an image, or a text to a text), as illustrated in Figure~\ref{fig:comparison}a. The model was shown to be very effective at so-called \textit{zero-shot learning}, which for CLIP means that the model can do image retrieval by ranking the similarity of images to a given label (such as ``a black cat"). This can be contrasted with traditional image classification, where the model is specifically trained to classify images into a predefined set of categories \shortcite<e.g.,>{Deng2009}. In addition to being more flexible (as it can use a virtually infinite set of categories), CLIP was also shown to be more robust against noise and variations in the images, compared to supervised image classification \shortcite{Radford2021}.

\begin{figure}[htb]
	\begin{center}
		\includegraphics[width=0.68\linewidth]{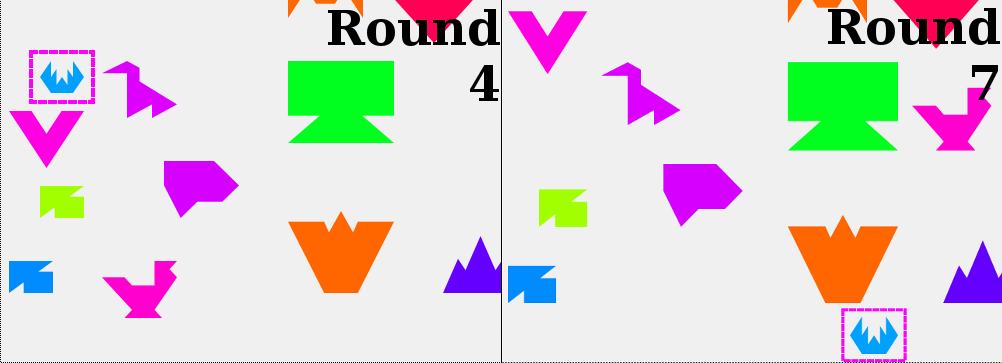}
		\begin{tabular}{| l l l |}
			\hline
			Speaker	& Round	&	Utterance \\
			\hline
			B	&	4	&	\textit{eh it looks like a blue crab sticking up his claws} \\
			\multicolumn{3}{| c |}{\ldots} \\
			A	&	7	&	\textit{it's the same the little crab again} \\
			\hline
		\end{tabular}
	\end{center}
	\caption{Example of repeated reference across rounds; the referent the participants have to collaboratively resolve is indicated here with a magenta outline (figure from \shortciteA{Shore2018-EMNLP}).}
	\label{fig:coreference}
\end{figure}

While such a model is potentially very useful for agents that need to ground language in vision, it is limited in that it is trained once, without any mechanism for updating its representations in light of new data, unless the entire model is retrained and the number of new examples is sufficient. This is clearly limiting the model's usefulness in real-life application scenarios for agents interacting in a dynamic environment. Not only will new objects with new properties emerge, but the way humans talk about objects changes over time. As has been shown repeatedly in experiments on human-human interaction, this is not only a long-term issue, but the exact meaning of language may often be negotiated and evolve during the course of a single interaction, and then develop into partner-specific language use \shortcite{Brennan1996,barr_anchoring_2002,Brennan2009,ibarra_flexibility_2016,Shore2018-EMNLP}. This phenomenon has been referred to as \textit{conceptual pacts} \shortcite{Brennan1996}, or more generally as \textit{alignment} in communication \shortcite{Pickering2006}. For example, if a hard-to-describe object is being referred to, the partners might establish a new name for it and then continue using that name for similar objects.  An example of this is shown in Figure~\ref{fig:coreference}, where two human subjects were asked to play a game where they take turns referring to tangram figures on a shared game board \shortcite{Shore2018-EMNLP}. In round 4, speaker B uses the referring expression ``a blue crab sticking up his claws", and in round 7, speaker A adopts the term ``crab" when referring to it again. Other pairs of subjects formed other conceptual pacts for the same shape, such as ``bat" or ``crown". Furthermore, humans can make use of \textit{semantic compositionality}, which means that if one person refers to an object with ``the blue crab", it is likely that the expression ``the pink crab" would be understood as a reference to a similarly shaped object in a different color. Thus, the novel language use ``crab" can be combined with the already established language use for colors. 

It is unreasonable to expect that a model like CLIP should be able to resolve such innovative language use, and it is not feasible to retrain the entire model frequently enough. Thus, if an artificial agent should be able to engage in such a task, it would clearly need to be able to adjust its language-to-vision grounding model, based on a single example. Even if we are not considering completely novel language use, there might be small misalignments between the agent's and the human's language use, or in their perceptual representations \shortcite{Chai2016}, which could lead to miscommunication if the model is not adapted. Often, the exact meaning of words depends largely on the context and the task at hand.

The ability to continually learn over time has been a long-standing challenge for machine learning and artificial intelligence, and this area of research has been referred to as \textit{continual} or \textit{lifelong learning} \shortcite{Parisi2019}. There are several problems involved in this. First, whereas humans can learn new concepts using only a few examples, machine learning models typically need several orders of magnitude more examples. Second, computational models have been shown to be prone to so-called \textit{catastrophic forgetting} \shortcite{Parisi2019}. Unless the model is re-trained entirely from scratch (which is infeasible for large models like CLIP), the updates to the model's parameters might interfere with previously learned knowledge, resulting in abrupt performance drops. This is also referred to as the \textit{stability-plasticity dilemma} \shortcite{Parisi2019}. 

Learning from few examples greatly depends on having powerful enough representations. Thus, the first problem has been addressed using \textit{transfer learning}, where a fixed \textit{foundation model} \shortcite{Bommasani2021} learns rich general representations from other (but related) tasks. This foundation model is then used as input to a simple classifier with only a few parameters (such as logistic regression), requiring only a few training examples. This is often referred to as \textit{few-shot learning} \shortcite{Wang2020}. Since the CLIP model is trained to learn powerful general representations, it was also shown to be fairly good as a foundation model for few-shot learning \shortcite{Radford2021}. In principle, an agent that sees a new object and hears it being referred to as ``a blue crab" by a human could train a few-shot classifier to be able to identify such objects in the future, as illustrated in Figure~\ref{fig:comparison}b. However, a problem with this form of few-shot learning is that it is based on the same principles as the conventional supervised image classification discussed above, where labels do not have any inherent meaning, but are instead treated as atomic symbols. If an agent using a language-image embedding model (such as CLIP) would learn a new label using this approach, it is not clear when it should use its foundation model to resolve language-image relationships (as in Figure~\ref{fig:comparison}a), and when it should apply the newly learned classifier for the specific label (as in Figure~\ref{fig:comparison}b). Moreover, those new categories would have no relationship to other (previously known or newly acquired) categories. For example, if the agent would learn how the term ``the red monitor" is used in a specific situation, it would not be able to infer that ``the red display" might be used in a similar way. In addition, it is unclear how it should be able to make use of semantic compositionality (as described above), i.e., to combine (in a principled way) the newly acquired language with the language it already knows in a compositional manner, to understand expressions such as ``the blue monitor".

In this paper, we propose \textbf{CoLLIE}, a simple, yet effective, model for \textbf{Co}ntinual learning of \textbf{L}anguage grounding from \textbf{L}anguage-\textbf{I}mage \textbf{E}mbeddings. The general principle of CoLLIE is illustrated in Figure~\ref{fig:comparison}c. Instead of learning a new model for each new concept (as in few-shot learning), the model relies on a foundation model of language-image embeddings in a joint embedding space (CLIP in our case), with zero-shot capabilities. We then use and update a separate \textit{transformation model} that makes adjustments to the language embedding to better fit the new concepts that are being learned, when needed. This transformation is done by predicting the \textit{difference vector} between the image and the language embeddings, and then adding this difference vector to the language embedding. This way, only the misaligned dimensions of the language embedding are corrected, while the others are unaffected. Since the continual learning is only taking place in this transformation model, and the foundation model is fixed, this is a very lightweight process. Our aim is to achieve the following characteristics: 

\begin{itemize}
    \item \textbf{Sample efficient}: We want to be able to learn new language-image mappings quickly with only a few examples. 
    \item \textbf{Computationally efficient}: The transformation model is very lightweight and relatively cheap to retrain. 
    \item \textbf{Generalizable}: We want to be able to use the newly learned concepts to understand new related concepts, and make use of semantic compositionality. 
    \item \textbf{Robust}: As the model learns new concepts, it should continue to perform equally well on tasks it could do before, and newly learned concepts should not interfere with each other. 
\end{itemize}

\section{Related Work}

Language grounding is a core problem of AI, and is related to the more general problem of symbol grounding. i.e., how the symbols used by an AI system get their meaning in terms of how they are anchored to the external world \shortcite{harnad1990}. The more specific problem of how language is grounded in vision has been addressed in different fields, including computer vision, computational linguistics, and artificial intelligence. In computer vision, the task of image classification has been studied extensively for a long time \shortcite<e.g.,>{Deng2009}. This is analogous to the problem of image retrieval, where images are ranked based on how well they match a certain class. In this formulation of the problem, each image is classified as belonging to a fixed number of classes. However, in real language use, language exhibits semantic compositionality and can express an almost infinite number of ``classes" by combining different concepts, such as ``the black cat on the mat". The problem of identifying the target (or referent) of such expressions has been referred to as \textit{referring expression comprehension} \shortcite{Qiao2020}. As discussed in Section~\ref{section-introduction}, this is typically done by encoding the image and text into a joint semantic embedding space, and rank the target referents according to how close they are in this space (ibid.). In (computational) linguistics, the problem of identifying referents of referring expressions in the external world is called \textit{exophoric} reference resolution, which is different from \textit{anaphoric} reference (or coreference) resolution, where references to entities in the past discourse are being resolved. 

The inverse problem, how to generate referring expressions based on the visual properties of the target referent, has also received considerable attention, earlier using rule-based approaches \shortcite{Krahmer2012}, and more recently using neural language generation conditioned on the visual encoding of the image \shortcite<e.g.,>{Panagiaris2021}. A challenge when generating referring expressions is that the model should ideally take the potential distractors into account (in order to uniquely identify the target), while at the same time being as efficient (brief) as possible (the Gricean Maxim of Quantity \shortcite{Grice1975LogicConversation}) \shortcite{Krahmer2012}. A related problem is that of image captioning, where the task is to describe an entire image or scene, rather than a specific object in an image \shortcite<e.g.,>{You2016}. 

When it comes to the visual grounding of language in interaction, there has also been a lot of research done in the areas of Visual Question-Answering (VQA) and multi-turn VQA, where visual language understanding and generation are combined and treated in an end-to-end fashion \shortcite{Kafle2017,Das2017}. Another field where this problem has been studied is human-robot interaction \shortcite<e.g.,>{Chai2016}. However, these studies typically assume that a fixed model of language grounding can be trained, and that the language use does not change after that. As discussed in Section~\ref{section-introduction}, the grounding of language is often (explicitly or implicitly) negotiated in dialog to handle new or specific situations; new words might be invented, or the exact meaning of words might change. 

One approach to accommodate partner-specific language use that evolves in dialog is to feed the dialog history as input to the model. An example of this is \shortciteA{Takmaz2020}, who trained a model to generate subsequent referring expressions, conditioned on past coreference chains in the dialog. However, this is only feasible for short-term effects, and not for language use that evolves over longer periods of time. For that, the parameters of the model need to be updated (i.e., some form of continual learning). \shortciteA{Shore2018-EMNLP} explored the accommodation of partner-specific language use in exophoric reference resolution for the game depicted in Figure~\ref{fig:coreference}. They showed that if the reference resolution model was retrained after each round (in light of new data), the performance increased significantly. While this was feasible to do given the small size of the model and the limited domain, it is clearly not feasible for real use case scenarios, where models like CLIP are trained on 400 million data points. For such cases, some form of continual learning is needed. 

Previous research on continual learning has mainly been done in the context of image classification, where there is a limited set of classes, but where new classes are gradually added to the model, so-called ``incremental class learning"  \shortcite<e.g.,>{rebuffi2017icarl,kemker2018fearnet,kemker2018measuring}. Early studies of continual learning in neural networks showed that the newly learned information interfered with previous knowledge in the shared representational resources, resulting in catastrophic forgetting \shortcite{mccloskey1989}. 
\shortciteA{Parisi2019} outline three basic approaches to alleviate catastrophic forgetting for continual learning in neural networks: First, various \textit{regularization approaches} may be used to impose constraints on the update of the model's parameters. 
Second, it is possible to allow the architecture of the network to change, for example, by adding neurons or layers. 
Third, \textit{complementary learning systems} are inspired by the human brain, in that they rely on an interplay between episodic memory (specific experience) and a semantic memory (general structured knowledge), where learning first happens in the former, and is eventually consolidated with the latter (during ``sleep"). One approach to avoid catastrophic forgetting is to store some of the older data points and mix these in when training with newer data points, a technique called ``rehearsal" \shortcite{Parisi2019}. A variant of this is ``pseudo-rehearsal", where a generative model is used to generate older data points \shortcite{kemker2018fearnet}. 

CoLLIE does not fit squarely into any of these three approaches, but comes closest to complementary learning in its use of a foundation model (where parameters are fixed) and a dynamic model (where learning happens). 
It should be noted that our work is a bit different from how continual learning is typically addressed and how the problem is typically formulated. In our case, we do not have a limited set of classes which is expanded during training. Instead, we assume a language-image embedding model (such as CLIP) that can be used for zero-shot reference resolution, and where the language can form a virtually endless number of ``classes" (i.e., the same way the problem is formulated in referring expression comprehension). Our task is then to adjust the model to learn a domain-specific \textit{language use}, while retaining the zero-shot performance of the model on the language use it was trained for. It should be stressed that this assumes that the model already has good representations of the visual domain, and the aim of the learning is not to improve those representations, but rather to learn how to better map those to new language use. Thus, the performance of CoLLIE is inherently constrained by the performance of the foundation model. 

While continual learning should ideally happen without any retraining/rehearsal and without keeping training data in memory, we accept keeping the newly learned examples in memory (and a small fixed set of negative examples, as we will see), since we do not have to keep the training data for the foundation model in memory. This way, only the transformation model needs to be re-trained using those new examples. 
It should also be noted that the parameters of the transformation model are relatively few and the footprint of the training samples is quite small (since we only store their embeddings). If we can achieve the sample efficiency objective, they should also be limited in quantity.

\section{Data, Task and Metric}
\label{sec:task}

For our evaluations, the task is that of image retrieval or referring expression comprehension \shortcite{Qiao2020}, i.e., to rank a set of candidate referents based on how well they match a referring expression. While referring expression comprehension is typically done on objects within images (ibid.), we use separate images for the object here, in order to more easily control the set of distractors and evaluate the performance based on that. As our metric, we use the Mean Reciprocal Rank (MRR), which is equal to 1 divided by the assigned rank of the correct candidate, yielding a score between 0 and 1. Thus, an MRR of 1 corresponds to ranking the correct candidate first and 0.5 corresponds to ranking it second (which can still be considered quite good if the number of candidates is large). The reason we choose MRR instead of accuracy is that it does not only take the top-ranked candidate into account, and therefore can be considered to be a more nuanced metric. Another metric that is sometimes used for similar tasks is Recall@K. However, we think that metric is better suited when there are several potential targets, which is not the case in our experiments. 

In this paper, we use two datasets. First, we use the \textbf{LAD dataset} (Large-scale Attribute Dataset) by \shortciteA{Zhao2018}, from which we selected a set of 200 categories belonging to the super-categories animals, fruits, electronics, and vehicles, with a total of 68,247 images. To verify CLIP's zero-shot performance on this dataset\footnote{For the experiments in this paper, we use the publicly released pre-trained CLIP model with the ViT-B/32 Vision Transformer architecture, unless otherwise stated (\url{https://github.com/openai/CLIP}).}, we did 20 iterations where we randomly selected one image per category (i.e., 200 images) and performed the ranking task using the LAD labels of the categories as referring expressions, yielding an MRR of 0.773. We think this confirms CLIP's impressive zero-shot performance on these types of images.

To study a more challenging set of images (for CLIP), we also use the images from the \textbf{KTH Tangrams dataset} \shortcite{Shore2018-LREC}, which were used for the task depicted in Figure~\ref{fig:coreference}. To assess CLIP's zero-shot performance on these tangram figures, we took the 17 shapes used in the study and picked a subset of five colors (red, green, blue, yellow, and purple), constituting a set of 85 candidate referents. The referring expressions were constructed by combining the color with the name of the shape used by the authors of the paper (e.g., ``the blue giraffe"). As expected, CLIP's zero-shot performance on these referents is not as good, only yielding an overall MRR of 0.310. (The MRR for the individual shapes are shown in Figure~\ref{fig:tangram_zero_shot}). While some shapes are identified correctly (``mountain", ``barn"), most of them are not. This is of course understandable, given that these images are not representative of CLIP's training data. 

\begin{figure}[h]
  \begin{center}
    \includegraphics[trim={1.5cm 0.2cm 0.2cm 0.2cm},clip,width=0.6\textwidth]{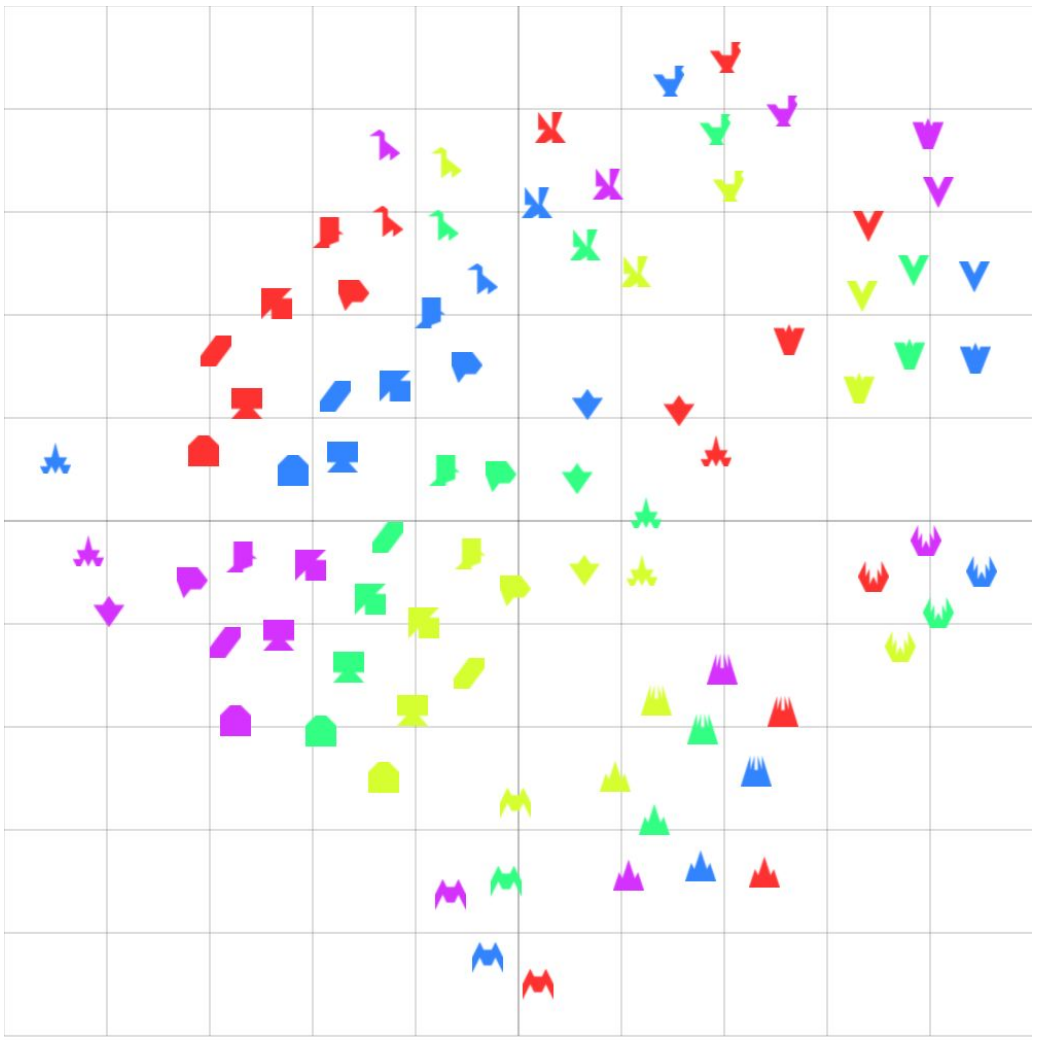}
  \end{center}
  \caption{t-SNE dimensionality reduction of the colored tangram CLIP embeddings.}
  \label{fig:tsne}
\end{figure}

In fact, it was not that easy for the human participants in the experiment to do this task either, at least not for their initial attempts. However, as discussed in Section~\ref{section-introduction}  and shown in Figure~\ref{fig:coreference}, they soon started to invent names for the different shapes, forming conceptual pacts after repeated interactions and making the interactions more efficient over time \shortcite{Shore2018-EMNLP}. If an artificial agent should be able to engage in such a task, it would clearly have to be able to apply some form of continual learning in the way outlined in the introduction. For this to work, CLIP still needs to have a good representation of the images. To investigate whether this is the case, we performed a t-SNE analysis \shortcite{VanDerMaaten2008} on the CLIP embeddings of the colored tangram images to reduce the 512 dimensions to 2 dimensions, as illustrated in Figure~\ref{fig:tsne}.
As can be seen, the shapes and colors seem to form clusters and to be handled in a somewhat consistent fashion, which suggests that it might be possible to associate them with names. We will discuss this task in more detail in Section~\ref{section-experiments}.

\section{The CoLLIE Transformation}

\begin{figure}
    \centering
    \begin{minipage}{0.55\textwidth}
        \centering
        \includegraphics[width=1.0\textwidth]{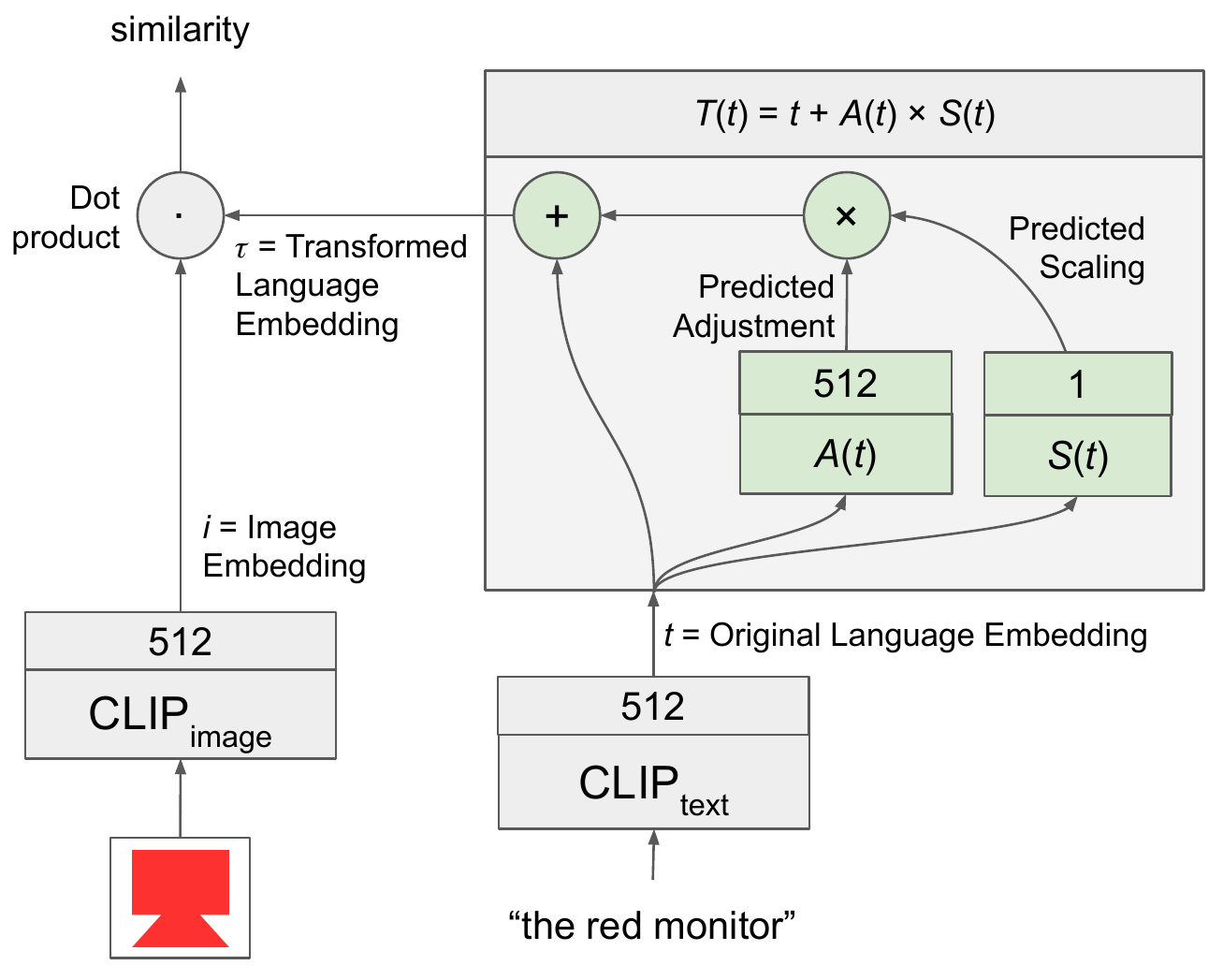}
        \caption{The CoLLIE transformation.}
        \label{fig:transformation}
    \end{minipage}\hfill
    \begin{minipage}{0.45\textwidth}
        \centering
        \includegraphics[width=1.0\textwidth]{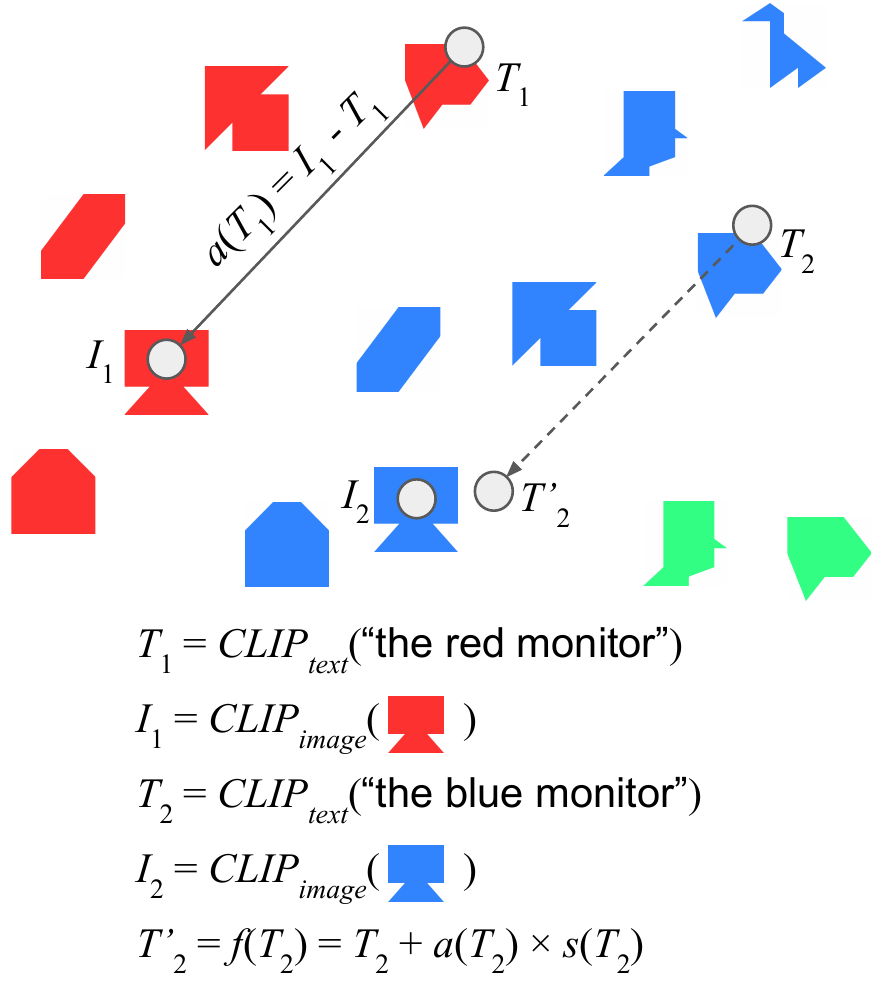} 
        \caption{A principled illustration of the intuition behind CoLLIE.}
        \label{fig:principle}
    \end{minipage}
\end{figure}

The idea behind CoLLIE is to learn a \textbf{transformation function}, \(T' = f(T) : \sR^{512} \rightarrow \sR^{512}\), which takes the CLIP embedding of the text, $T$, and returns another transformed embedding, $T'$, that better represents the new language use, and is closer to the CLIP image embedding $I$, as illustrated in Figure~\ref{fig:transformation}. It is important to note that in order to retain the zero-shot performance,  $f$ should in most cases return a similar output as input, unless the text has a domain-specific meaning that the model should correct for. 

The transformation function is modelled as \(f(T)=T+a(T)\times s(T)\), where \(a(T) : \sR^{512} \rightarrow \sR^{512}\) is an \textbf{adjustment function}, and \(s(T) : \sR^{512} \rightarrow [0,1]\) is a \textbf{scaling function}. As we will see, this scaling function helps to retain the zero-shot performance of the model. It can be noted that this principle is similar to that of residual connections in neural networks \shortcite{he2016deep} and Gated Linear Units \shortcite{Dauphin2017}. 

Training examples are stored as pairs of text and image embeddings \(\langle T,I \rangle\), and thus have a limited footprint (512+512 floats). \(a(T)\) is then trained to predict the \textit{difference vector}, \(I-T\), using the accumulated training examples. When doing the adjustment, the predicted difference vector is then added to $T$. The reason for predicting the difference vector (rather than $I$ directly) is that we want to learn which dimensions need to be corrected, while not affecting the dimensions that are already correct. We learn \(A\) using linear regression: \(a(T) = \beta T + m, \beta \in \sR^{512 \times 512}, m \in \sR^{512}\). To avoid overfitting (given the limited number of training examples) we use ridge regression (L2 regularization with \(\lambda=0.001\)). We will return to this and evaluate alternatives in Section~\ref{section-experiments}. 

The objective of the scaling function, $s$, is to return a value close to 1 when the input is a text that should be transformed (i.e., close to any example in the training set), and close to 0 otherwise. 
We learn $s$ using a regression model, where the accumulated training examples are used as positive examples (with training target 1). As negative examples (with training target 0), we simply use a list of the 1,000 most common nouns in English (representing expressions that should not be transformed). The rationale for using nouns is that all referring expressions are expected to have at least one head noun (since they are noun phrases), and that these common nouns should cover the embedding space fairly well. For example, the embedding for the word ``shoe" should be fairly similar to ``the large shoe" or ``the shoe with laces". As we add positive training examples, we create exceptions for nouns (and noun phrases) that should be adjusted. For our initial tests, we learn $s$ using support vector regression (SVR) with a linear kernel (coerced in the range \([0,1]\)), but we will evaluate alternative models in Section~\ref{section-experiments}.

Figure~\ref{fig:principle} illustrates the intuition behind the model: Given that we have a reference to an image (``the red monitor"), we encode it using CLIP and get an embedding \(T_1\). As can be seen, in this case, the text embedding is not very close to the embedding of the corresponding image \(I_1\), and will thus retrieve the wrong referent. To teach the model to make better predictions in the future, we add the pair \(\langle T_1,I_1 \rangle\) as a training example for $a$ and \(\langle T_1,1 \rangle\) as a training example for $s$. Using the accumulated training examples, we train \(a\) to approximate the difference vector between the embedding of the image and the text (\(I_1 - T_1\)), and $s$ to predict a value close to 1 for the input \(T_1\). Recall that we retrain only the added transformation and scaling functions, while the CLIP weights remain frozen. Now, when a new referring expression is to be resolved, ``the blue monitor", the expression is encoded by CLIP into \(T_2\). Again, directly using this embedding would result in a poor match for this domain. If we now apply the learned adjustment function \(a(T_2)\), it is likely to return a similar vector as \(I_1 - T_1\) (given that \(T_2\) is relatively close to \(T_1\) and that we do not have any other, more similar, training examples). Similarly, \(s(T_2)\) is likely to return a value fairly close to 1 (given that \(T_2\) is fairly close to \(T_1\)). We now get a new vector \(T'_2 = T_2 + a(T_2) \times s(T_2)\), which is indeed closer to the true referent \(I_2\). Thus, while the color dimensions seem to align well between \(T_2\) and \(I_2\), the predicted adjustment (the difference vector) corrects for the misaligned shape dimensions. 


\section{Experiments}
\label{section-experiments}
To evaluate CoLLIE, we devised two experiments. In the first experiment, we test whether the model can learn novel pseudo-words for already known objects (the LAD dataset). The focus of this experiment is to ensure that the model can generalize to other images of the same object and that the model can retain its zero-shot performance. In the second experiment, we test whether the model can learn to ground known words to novel objects (the KTH Tangrams dataset). Here, we want to ensure that the model can utilize semantic compositionality and generalize to similar language use.

\subsection{Experiment I: Learning Pseudo-words for Realistic Objects}
\label{sec:experiment-lad}

\begin{figure}[t]
\begin{center}
\includegraphics[width=1.0\textwidth]{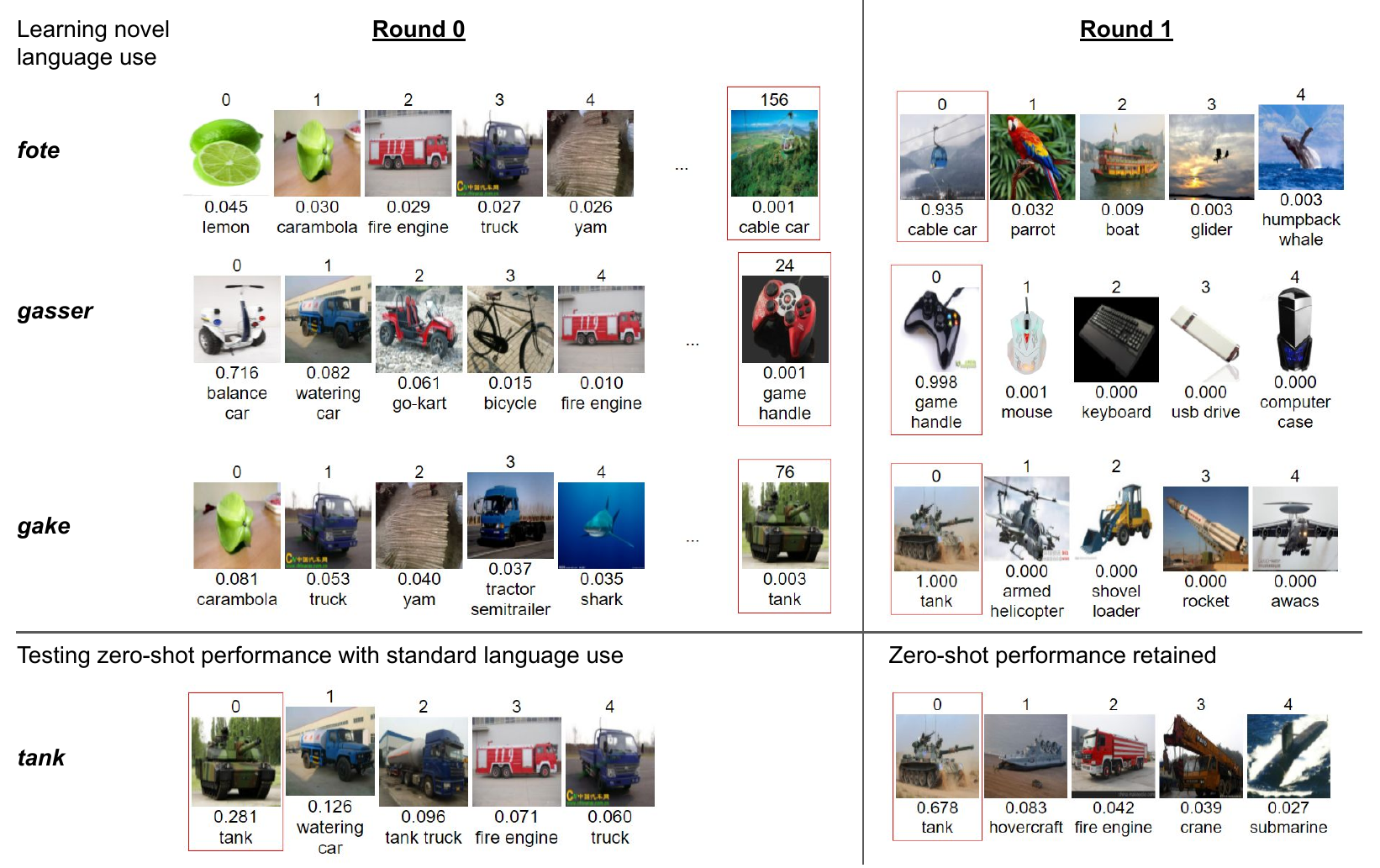}
\end{center}
\caption{An example from the first two rounds of Experiment I. The referring expression is shown to the left and the correct target is marked with a red square. At round 0 (before any training), CoLLIE obviously doesn't recognize the new words, and they are ranked very low. At round 1 (after having seen just one example of each new word), it correctly ranks them first. The zero-shot performance of the model (identifying the referents of the original names) is unaffected.}
\label{fig:lad-example-training}
\end{figure}

We first devised an evaluation scheme to see whether the model can learn new words for photographic images from the LAD dataset, while we monitor the retained zero-shot performance during training (i.e., to make sure that the model still understands the existing words). 
We randomly select a set of \(N\) categories, \(C_{train}\), (out of the 200 categories) for which we want to teach the model new names. We then assign a new name for each of these \(N\) categories, using randomly selected pseudo-words from the Novel Object and Unusual Name (NOUN) Database (``boskot", ``derd", ``tust", etc.) \shortcite{Horst2016}. Training is then performed over five and testing is performed over six \textit{rounds} (the initial round of testing is performed without training, which reflects the model's zero-shot performance). At the beginning of each round, we randomly select one image for each of the 200 categories, without ever reusing images between rounds. We then let the model rank the 200 images as potential referents for each pseudo-word, and the MRR is computed as described in Section~\ref{sec:task}. At the end of each round, we add the images from \(C_{train}\) and their associated pseudo-words as training examples (i.e., one example per category) to the model, and retrain it. An example of the first two rounds is shown in Figure~\ref{fig:lad-example-training}. This whole procedure is repeated over 50 \textit{iterations} (with new pseudo-words and categories randomly selected and assigned), in order to get a smooth average performance per round. 

We evaluate and compare the performance using (1) the CoLLIE model,  (2) the fixed CLIP model, and (3) a few-shot classifier based on logistic regression \shortcite<implemented in  the same way as in>{Radford2021}. For the few-shot classifier, each pseudo-word is treated as a class. To study the effect of the scaling function, we also add (4) the CoLLIE model without the scaling function.

\begin{figure}[t]
\begin{center}
\includegraphics[width=1.0\textwidth]{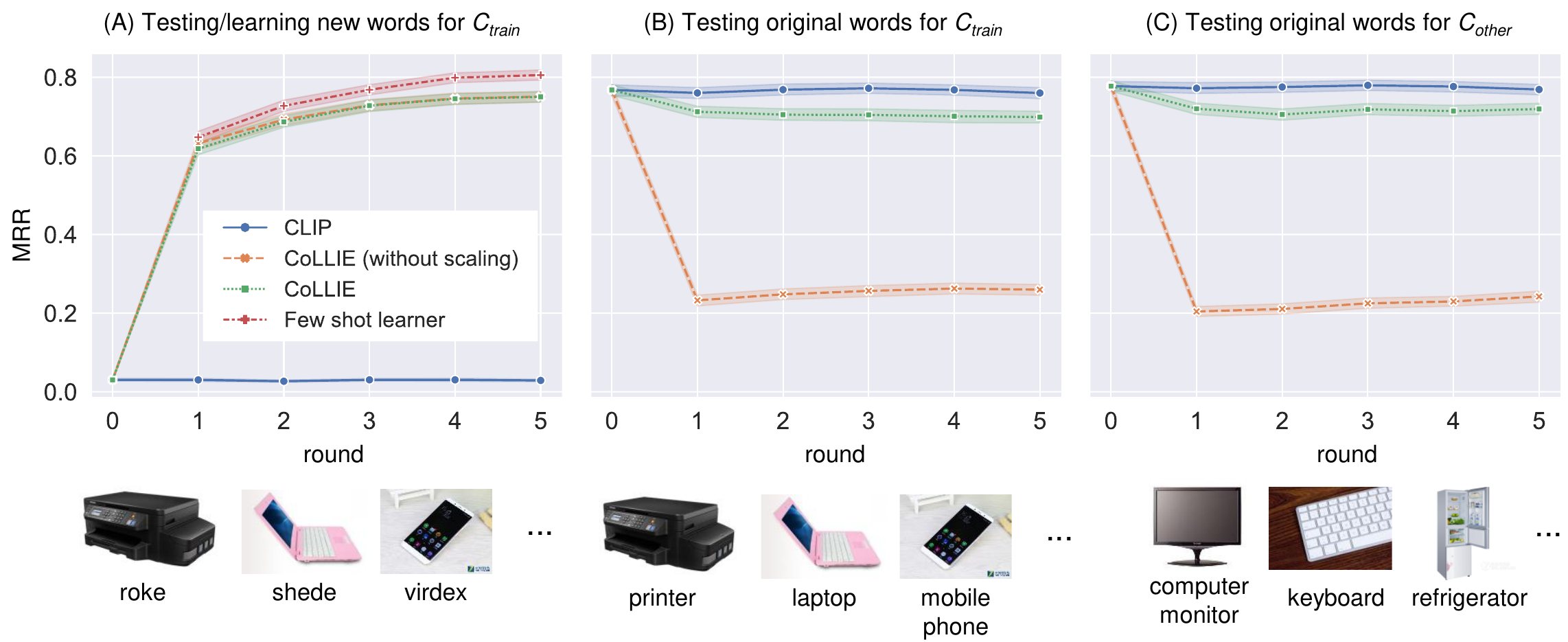}
\end{center}
\vspace{-0.3cm}
\caption{Performance of CoLLIE (with SVR scaling function) over five rounds of training on the LAD dataset (averaged over 50 iterations, 95\% CI), where \(|C_{train}|=50\). One training example per category is added per round. (A) shows continual learning performance. (B) and (C) show retention performance. Few-shot learner is only applicable for pane (A). }
\vspace{-0.3cm}
\label{fig:lad-training}
\end{figure}

The results are shown in Figure~\ref{fig:lad-training}(A), where \(N=50\). As can be seen, CoLLIE quickly learns the new pseudo-words, and reaches a fairly good performance (0.618) already after one round (i.e., when it has only been provided with one example per category), increasing to 0.750 at the final round (where five examples have been provided), which is quite close to CLIP's zero-shot performance of 0.773 for the original words on this dataset. Here, the scaling function has very little effect. Since the CLIP model is not doing any learning, it obviously has a very poor zero-shot performance on these new words. However, the few-shot classifier has a slightly better performance than CoLLIE, especially after five examples are added (0.805). This is perhaps not very surprising, given that it is optimizing this classification task, rather than transforming the embedding space. Also, in this experiment, CoLLIE does not benefit from generalization of the learned words, as they are arbitrarily assigned and there is no semantic compositionality effect (which we will get back to in Experiment II).

To study the retained zero-shot performance of the model during training, we also plot the performance of the models when using the original words for the 50 categories in \(C_{train}\), in Figure~\ref{fig:lad-training}(B). As can be seen, the original names for those categories can still be resolved by CoLLIE with a slight (but not catastrophic) drop in performance compared to the static CLIP model (0.699 vs. 0.760 at the final round), even though the model has also learned new words for them. 
In this case, the scaling function is important and without it, the performance drops considerably (to 0.260). Similarly, for each round, we also study the models' retained zero-shot performance on 50 randomly selected categories, \(C_{other}\), which were not part of \(C_{train}\), using their original names. This is shown in Figure~\ref{fig:lad-training}(C). Again, when using the scaling function, CoLLIE does not seem to interfere much with CLIP's original zero-shot performance (0.719 vs. 0.769 at the final round), while there is a drastic drop in performance (to 0.242) when the scaling function is not used. Since the few-shot learner cannot be applied to these problems, its performance is not plotted in pane B-C. 

\subsubsection{Effects of Scaling and Adjustment Functions}

As these results show, the scaling function is important for the retention of the model's zero-shot performance. As mentioned earlier, we initially used SVR (with a linear kernel) for the scaling function, as it was showing promising performance. In Table~\ref{table:results-scaling}, we also investigate different implementations of the scaling function\footnote{We use the implementations in scikit-learn \shortcite{scikit-learn} with standard parameters. The standard error of the mean is consistently around 0.007, and is therefore not reported in the table.}: using different SVR kernels (RBF, sigmoid and linear), linear regression, logistic regression, and a KNN regressor (with a weighted distance function and \(K=10\)). As can be seen, when taking both continual learning and retention performance into account, SVR (with either RBF or linear kernel) compares favorably compared to logistic and linear regression. The KNN regressor also shows a good performance. One explanation why KNN and SVR classifiers might perform better is that they both give more weight to the nearest neighbors, and handle class imbalance better (between the newly learned words and the negative examples). For a good scaling function, we want it to output a score close to 1 when the expression is part of the newly learned words, and close to 0 otherwise. For this, the closest words in the training set should be the most informative ones. 

\begin{table}[t]
    \caption{Performance of the model (MRR) on the LAD dataset (averaged over 50 iterations), with different implementations of the scaling function, where \(|C_{train}|=50\).}
    \label{table:results-scaling}
    \begin{center}
\begin{tabular}{p{4.7cm}|rr|rr|rr}

\hline
&\multicolumn{2}{p{3cm}|}{Learning new words for \(C_{train}\)} 
& \multicolumn{2}{p{3cm}|}{Testing original words for \(C_{train}\)} 
& \multicolumn{2}{p{3cm}}{Testing original words for \(C_{other}\)} \\
\hline
Round & 1 & 5 & 1 & 5 & 1 & 5 \\ 
\hline
SVR (RBF)           & 0.639 & 0.743 & 0.731 & 0.739 & 0.711 & 0.722 \\ 
SVR (sigmoid)       & 0.582 & 0.672 & \textbf{0.764} & \textbf{0.765} & \textbf{0.749} & \textbf{0.752} \\ 
SVR (linear)        & 0.627 & 0.723 & 0.743 & 0.749 & 0.726 & 0.732 \\ 
Logistic regression & 0.633 & \textbf{0.756} & 0.656 & 0.679 & 0.649 & 0.657 \\ 
Linear regression   & 0.648 & 0.755 & 0.659 & 0.674 & 0.636 & 0.656 \\ 
KNN (K=10)          & \textbf{0.650} & 0.755 & 0.762 & 0.763 & 0.744 & 0.748  \\
\hline
No scaling          & \textbf{0.650} & 0.755 & 0.240 & 0.270 & 0.210 & 0.243 \\
\hline
No scaling, negative examples
in adjustment function  & 0.541 & 0.728 & 0.563 & 0.416 & 0.552 & 0.391 \\ 
\hline

\end{tabular}
\end{center}
\end{table}

A potential alternative to a separate scaling function is to instead use the negative examples (common nouns) as training examples for the adjustment function, with zero-length vectors as targets. This could potentially allow the adjustment function to directly learn that no adjustment should be applied for those examples. This is shown in the last row of Table~\ref{table:results-scaling}. As can be seen, the performance is much lower (especially when it comes to retaining zero-shot performance), which motivates the need for a separate scaling function.

\begin{table}[t]
    \caption{Performance of the model (MRR) on the LAD dataset (averaged over 50 iterations), with different implementations of the adjustment function, where \(|C_{train}|=50\).}
    \label{table:results-adjustments}
    \begin{center}
\begin{tabular}{p{4cm}|rr|rr|rr}

\hline
&\multicolumn{2}{p{3cm}|}{Learning new words for \(C_{train}\)} 
& \multicolumn{2}{p{3cm}|}{Testing original words for \(C_{train}\)} 
& \multicolumn{2}{p{3cm}}{Testing original words for \(C_{other}\)} \\
\hline
Round & 1 & 5 & 1 & 5 & 1 & 5 \\ 
\hline
Linear        & \textbf{0.640} & \textbf{0.744} & 0.729 & 0.148 & 0.708 & 0.0825 \\
Ridge (\(\lambda=0.0001\)) & \textbf{0.640} & \textbf{0.744} & 0.730 & 0.738 & 0.708 & 0.722 \\
Ridge (\(\lambda=0.001\))  & \textbf{0.640} & 0.743 & 0.731 & 0.739 & 0.711 & 0.722 \\
Ridge (\(\lambda=0.01\))   & 0.625 & 0.740 & 0.741 & 0.740 & 0.724 & 0.726 \\
Ridge (\(\lambda=0.1\))    & 0.474 & 0.702 & \textbf{0.758} & \textbf{0.751} & \textbf{0.743} & \textbf{0.740} \\
\hline

\end{tabular}
\end{center}
\end{table}

In Table~\ref{table:results-adjustments}, we show a similar comparison, but with different variants of the adjustment function (using SVR with RBF kernel for the scaling function). For the continual learning of the new words, both linear and ridge regression have similar performance, as long as the regularization in the ridge regression is not too strong (i.e., \(\lambda\) is set too high). For the retention performance (testing existing words), the simple linear regression results in catastrophic forgetting as more training examples are added (see rightmost column), while the ridge regression keeps the performance at an acceptable level. Thus, we conclude that ridge regression with \(\lambda=0.001\) gives a fairly good trade-off. 

\subsubsection{Effects of Training Size}

To further investigate the performance, we also run experiments with different numbers of classes/pseudo-words to learn (\(N = |C_{train}|\)), and different numbers of negative examples in the scaling function. The results are shown in Table~\ref{table:results-lad}. As can be seen in (A), the continual learning performance is relatively stable for different values of $N$, and the choice of scaling function or number of negative examples has no large impact. However, as seen in (B), the zero-shot retention performance is clearly affected as $N$ increases. This is especially true when only 100 negative examples are used for the scaling function. Thus, it is possible that the drop in retention performance could be mitigated by adding even more negative examples, as $N$ increases. As the number of pseudo-words to learn increases, the KNN regressor also shows a  clearly better performance than SVR, in terms of retention. Investigating suitable scaling functions for large values of $N$ is an important topic for future work. 

\begin{table}
    \caption{Performance of the model (MRR) on the LAD dataset (averaged over 50 iterations), with different implementations of the scaling function, number of negative examples (n/e), and number of classes/pseudo-words to learn (\(N\)). (A) shows continual learning performance and (B) shows retention performance.}
    \label{table:results-lad}
    \begin{center}
\begin{tabular}{l|rrrr|rrrr}
\hline

\(N=|C_{train}|\) & 10 & 50 & 100 & 150 & 10 & 50 & 100 & 150 \\ 
\hline
\textbf{(A) New words} &\multicolumn{4}{l|}{\(C_{train}\): After 1 example} 
& \multicolumn{4}{l}{\(C_{train}\): After 5 examples} \\
\hline
SVR (1,000 n/e) & 0.616 & 0.618 & 0.636 & 0.634 & 0.770 & 0.750 & 0.746 & 0.751 \\ 
KNN (1,000 n/e) & 0.646 & \textbf{0.648} & 0.637 & 0.635 & 0.738 & \textbf{0.758} & \textbf{0.749} & 0.747 \\
SVR (100 n/e) & 0.639 & 0.629 & 0.639 & 0.636 & 0.770 & 0.750 & 0.746 & 0.752 \\ 
No scaling & \textbf{0.654} & 0.632 & \textbf{0.641} & \textbf{0.637} & \textbf{0.774} & 0.751 & 0.746 &\textbf{ 0.753} \\ 
CLIP baseline & 0.033 & 0.030 & 0.030 & 0.028 & 0.028 & 0.029 & 0.030 & 0.031 \\ 
\hline
\textbf{(B) Orig. words} & \multicolumn{4}{l|}{\(C_{train}\): After 5 examples} 
& \multicolumn{4}{l}{\(C_{other}\): After 5 examples} \\ 
\hline
SVR (1,000 n/e) & 0.780 & 0.753 & 0.633 & 0.529 & 0.774 & 0.725 & 0.645 & 0.522 \\
KNN (1,000 n/e) & 0.778 & \textbf{0.767} & 0.707 & 0.641 & \textbf{0.779} & 0.743 & 0.707 & 0.631 \\ 
SVR (100 n/e) & 0.769 & 0.621 & 0.450 & 0.333 & 0.732 & 0.638 & 0.443 & 0.314 \\ 
No scaling & 0.515 & 0.260 & 0.127 & 0.078 & 0.361 & 0.242 & 0.112 & 0.054 \\ 
CLIP baseline & \textbf{0.798} & 0.760 & \textbf{0.768} & \textbf{0.773} & \textbf{0.779} & \textbf{0.769} & \textbf{0.783} & \textbf{0.774} \\ 
\hline

\end{tabular}
\end{center}
\end{table}

\subsection{Experiment II: Learning Language for Tangram Figures}
\label{sec:experiment-tangrams}

The biggest expected benefit of CoLLIE comes from its ability to generalize from the language it is learning, for example through semantic compositionality, which was not addressed in Experiment I. We thus devised an evaluation scheme to see how quickly the model can learn to identify the colored tangram shapes (introduced in Section~\ref{sec:task}), for which it clearly had a very poor zero-shot performance. Here, we expect the model to benefit from the compositionality of the referring expressions. As discussed earlier, given that it has learned what visual properties to associate with the phrase ``the blue rock", it should be able to generalize this understanding to ``the red rock". 

\begin{figure}[t]
\begin{center}
\includegraphics[width=1.0\textwidth]{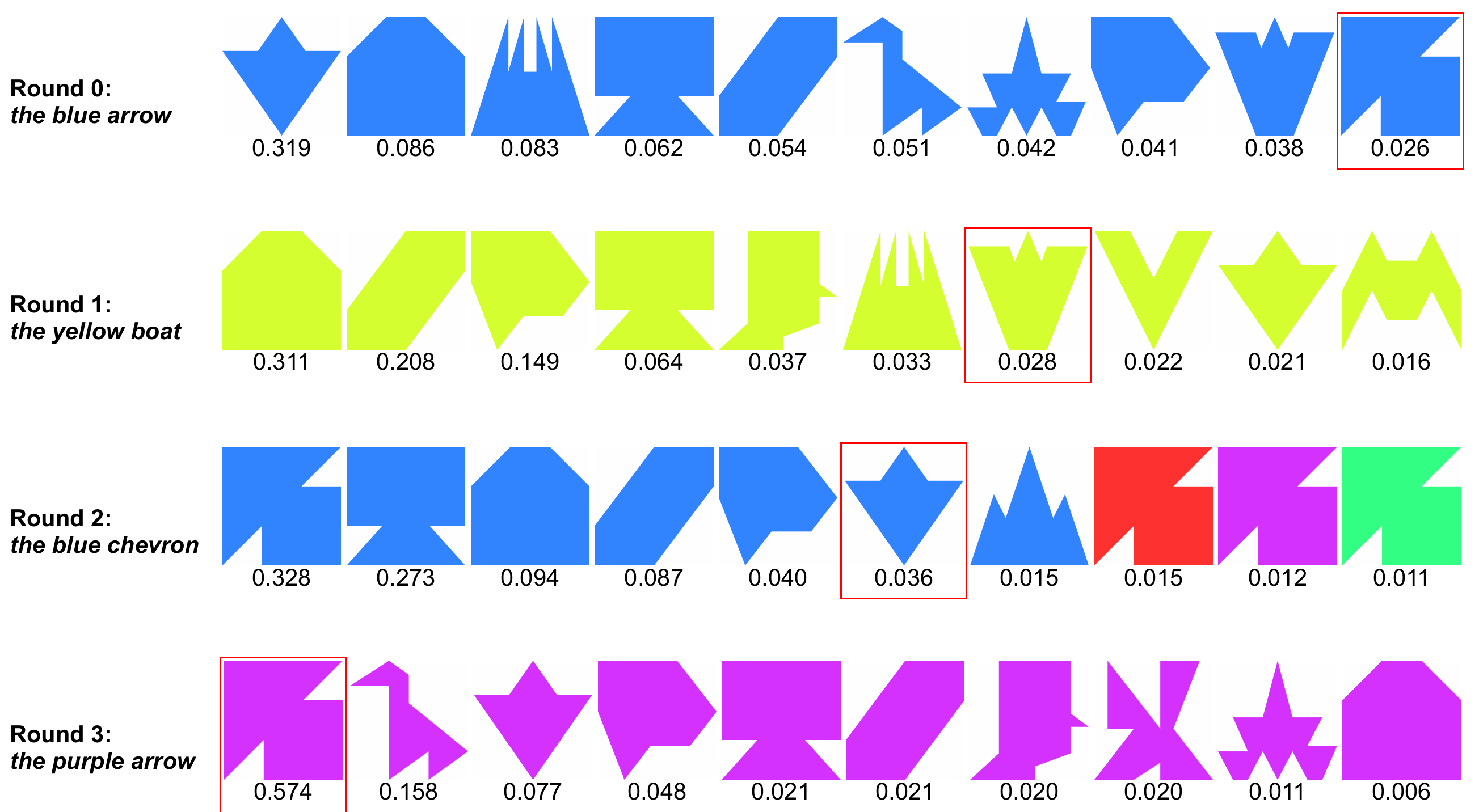}
\end{center}
\caption{An example of the first four rounds of one iteration of Experiment II. Each round shows the randomly selected referring expression, the top 10 candidates as ranked by the model left-to-right (with their softmax scores), as well as the correct referent (marked with a red square). In round 0--2, new names of shapes are introduced, but after having seen one example of a ``blue arrow" in round 0, the model correctly identifies the ``purple arrow" in round 3.}
\label{fig:tangrams-example-training}
\end{figure}

Again, the task is to rank the 85 potential referents, given a referring expression. Similar to Experiment I, the model starts out with no training examples (round 0). We then train the model over 30 \textit{rounds}. In each round, one random referent is picked, the model's performance (in terms of MRR) on this referent is assessed, the image-text pair of the referent is added to the training set, the model is retrained, and a new round begins.  This whole procedure is then repeated over 3,000 \textit{iterations}, resetting the model after each iteration. An example of the first few rounds in one iteration is shown in Figure~\ref{fig:tangrams-example-training}. 

\begin{figure}[t]
\begin{center}
\includegraphics[width=0.8\textwidth]{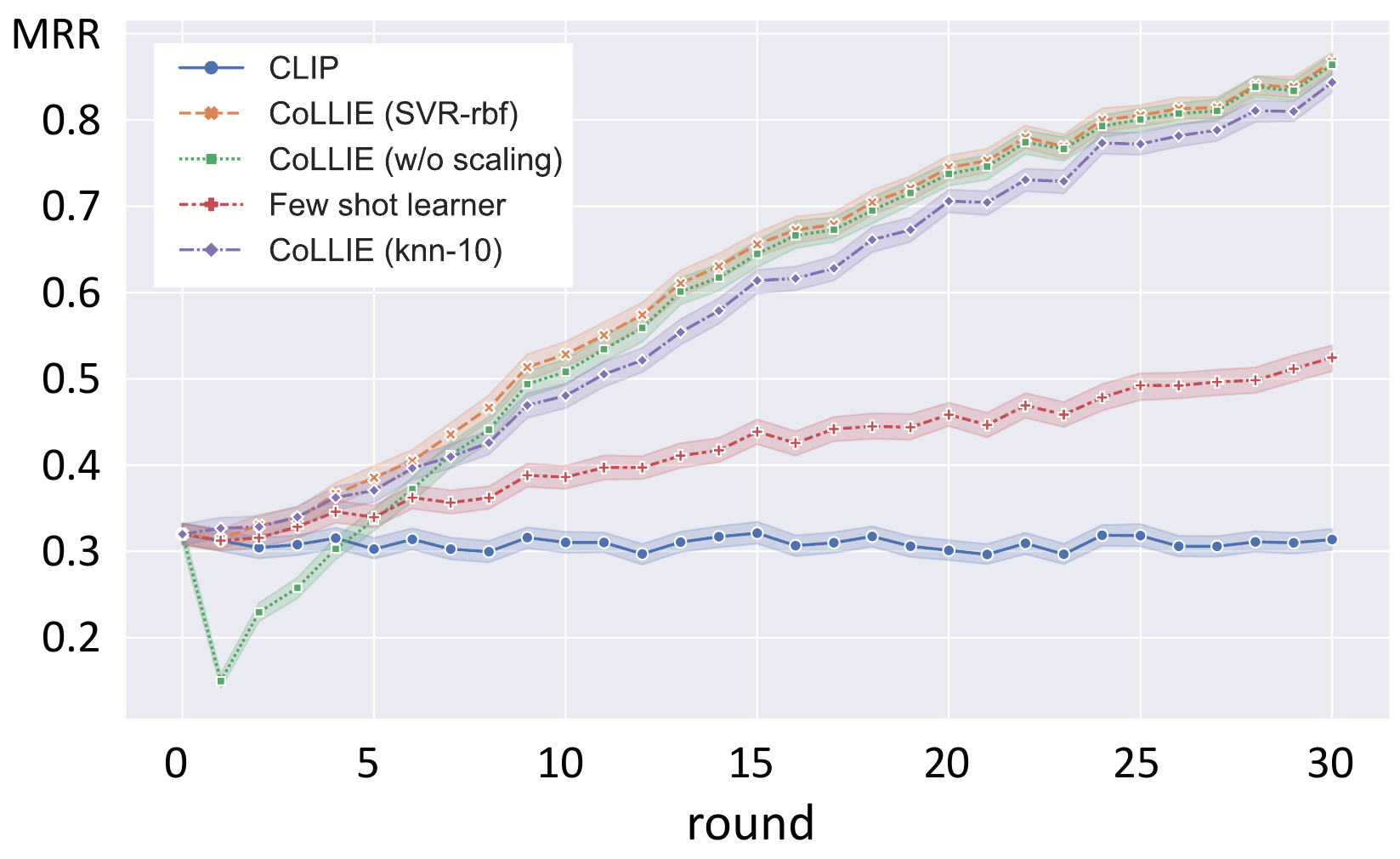}
\end{center}
\caption{Performance of the model over 30 rounds of training (averaged over 3,000 iterations) when training on the colored tangrams (95\% CI). One new training example is added per round. }
\label{fig:tangrams-training}
\end{figure}

The MRR per round (over all iterations) is illustrated in Figure~\ref{fig:tangrams-training}. We do a similar comparison with other models as in Experiment I. Here, we let the few-shot learner (again, a logistic regression classifier) fall back on CLIP when it is presented with a referring expression it has not seen before. We then add the new referring expression as a new class for the few-shot learner, and retrain it.

As can be seen, CoLLIE (with SVR scaling function) quickly learns the names for the tangrams, reaching an MRR of 0.860 after 30 rounds. Note that the 85 images are all unique in terms of shape-color combinations, which means that the model must be able to generalize in order to achieve this performance. In contrast, the few-shot learner  has much worse performance, as new referring expressions (classes) are introduced in most rounds in the beginning (unless the exact same object happened to be picked twice), and it has no way of generalizing from already learned classes.

This supports the hypothesis that CoLLIE should be able to benefit from the compositionality of language: After being taught what a ``red giraffe" looks like, CoLLIE is now better at identifying a ``blue giraffe", combining the base representation of ``blue" with the learned meaning of ``giraffe". 
To further confirm this ability, we also performed an experiment where we first train the model on all 17 shapes of one random color, and then evaluate it on the same shapes with different random colors. This was iterated 100 times. Whereas the CLIP baseline model (and the few-shot learner, which has to fall back on the CLIP model) only had an average MRR of 0.317 on these unseen combinations, CoLLIE achieved an MRR of 0.857. 

The intuition behind why this works was illustrated in Figure~\ref{fig:principle}: CoLLIE learns to predict the \textit{difference vector} that needs to be applied. Thus, if the color dimensions in the CLIP embedding were already aligned between the language and the image, there will not be any need to adjust those dimensions -- it is only the dimensions related to the shape that need to be adjusted. The fact that this works despite CLIP's representation being entirely distributed is interesting. 
The steady improvement also indicates that the learning of each concept does not interfere with the learning of other concepts. However, as can be seen in Figure~\ref{fig:tangrams-training}, without the scaling function, the model has a drastic drop in performance for the first rounds, which is likely because the newly learned adjustments are added too generously to unrelated referring expressions. 

Figure~\ref{fig:tangrams-training} also shows a comparison between the SVR and the KNN scaling functions. Unlike in Experiment I, SVR here gives a slightly better performance than KNN. This can perhaps be explained by the fact that, in this experiment, the scaling function has to give a high score to similar, but not identical, referring expressions, in order to generalize the adjustments it has learned. 

\begin{figure}[ht]
\includegraphics[width=1.0\textwidth]{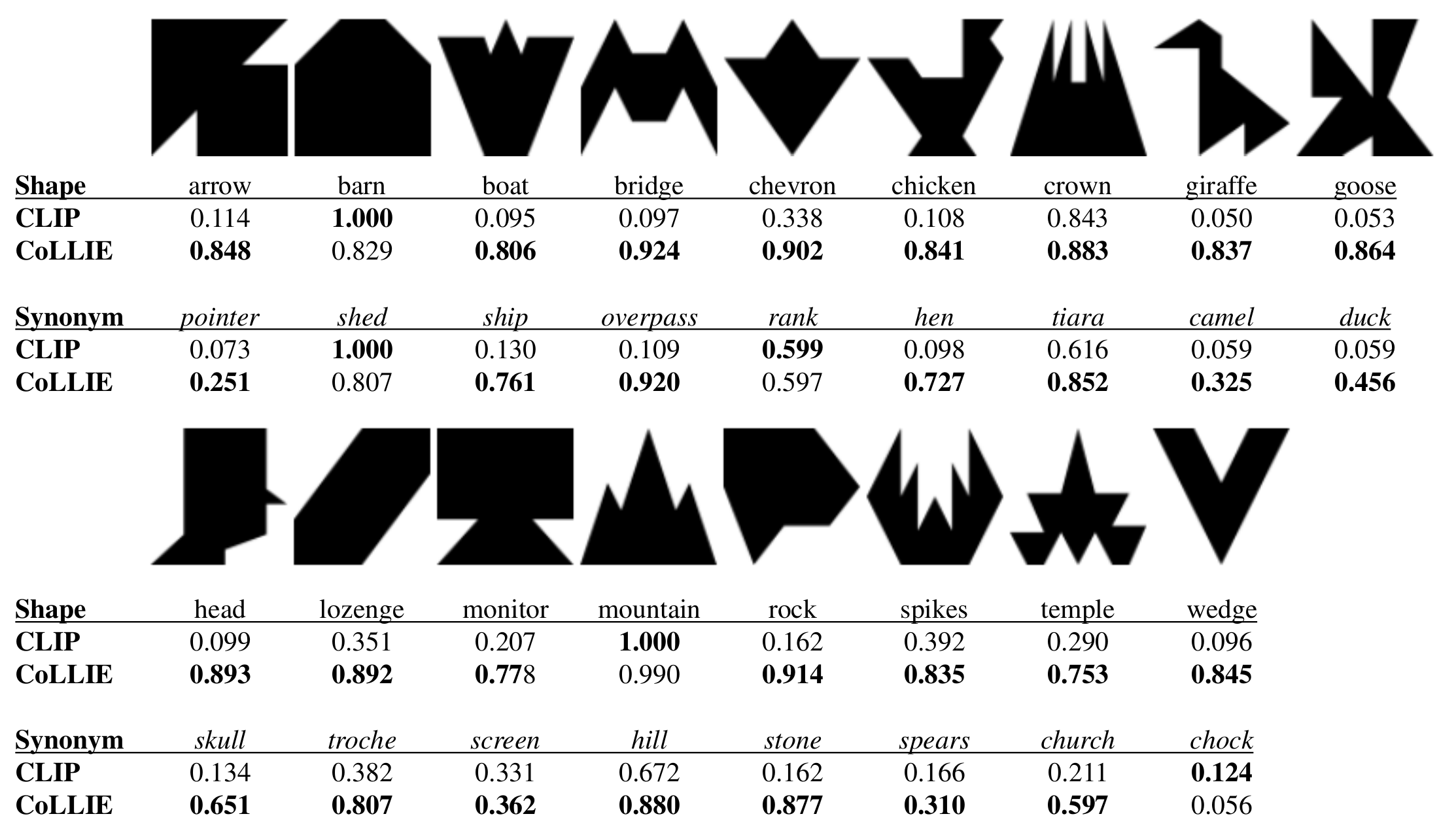} 
\caption{Performance (MRR) on individual tangram shapes and their synonyms. Both the original zero-shot performance of CLIP (over the 85 candidates), and the performance of CoLLIE (after training on 30 examples with the original names), are shown.}
\label{fig:tangram_zero_shot}
\end{figure}

\subsubsection{Generalizing with Synonyms}

As a further (limited) test to verify the model's ability to generalize, we substituted the names of the tangrams with synonyms\footnote{Taken from \url{thesaurus.com}} (``barn"\(\rightarrow\)``shed", ``chicken"\(\rightarrow\)``hen", etc.). This way, we formed referring expressions such as ``the blue hen". Using the CoLLIE model trained for 30 rounds as described above, we then evaluated these expressions (over all 3,000 iterations). The MRR for these was 0.602, which is clearly better than the baseline of 0.290 (using CLIP directly), providing further evidence for the model's ability to generalize. Given that many of the names had no obvious synonyms, their individual performance varied greatly (MRR 0.056-0.920). A breakdown of these results can be found in Figure~\ref{fig:tangram_zero_shot}. 

\subsubsection{Effects of Different Vision Backbones}

\begin{table}[t]
    \caption{Performance of CoLLIE (with SVR scaling function) on Experiment II (tangram figures), with different vision backbones in the CLIP model. Average results (MRR) over 3,000 iterations are shown. The rightmost column shows the number of dimensions in the CLIP embeddings.}
    \label{table:results-vision}
    \begin{center}
\begin{tabular}{l|rrrrr|r}

\hline
Round & 0 (CLIP) & 1 & 5 & 20 & 30 & CLIP Dim.\\ 
\hline
ResNet-50        & 0.300 & 0.269 & 0.355 & 0.630 & 0.848 & 1024\\
ResNet-101       & 0.293 & 0.305 & 0.407 & 0.680 & 0.860 & 512\\ 
ResNet-50x4      & 0.237 & 0.269 & 0.367 & 0.658 & 0.849 & 640\\ 
ViT-B/32         & 0.321 & 0.307 & 0.386 & 0.661 & 0.857 & 512\\ 
\hline

\end{tabular}
\end{center}
\end{table}

So far, we have used the CLIP model with the ViT-B/32 vision transformer architecture to encode images. To test whether CoLLIE can also work with different variants of the foundation model, we did an experiment where we used different image encoders in the CLIP model (all part of the public release). Not only do these different models have different vision architectures, they also produce text and image embeddings with different dimensionalities. Apart from the vision transformer, there are also three variants of ResNet \shortcite{he2016deep}: ResNet-50, ResNet-101, and an EfficientNet-style model scaling using approximately 4x the compute of a ResNet-50 (ResNet-50x4). The performance on the tangram task with these different models is shown in Table~\ref{table:results-vision}. As can be seen, although the CLIP zero-shot performance with the ResNet-50x4 model is lower, the performance of CoLLIE seems to be relatively unaffected by the choice of vision backbone. 

\section{Discussion}

Whereas most previous studies on continual learning have focused on incremental class learning in image classification \shortcite<e.g.,>{rebuffi2017icarl,kemker2018fearnet,kemker2018measuring,Benavides2020}, we have addressed a somewhat different problem. In our case, we rely on a foundation model capable of zero-shot image retrieval, where there is not a finite set of classes to learn. Instead, such a model encodes the image and language into a joint embedding space, in order to assess how well they match semantically. The advantage of such an approach is that we can resolve a virtually endless number of referring expressions, exhibiting semantic compositionality, such as ``the red barn". This is also the approach that is typically taken to address the problem of referring expression comprehension \shortcite{Qiao2020}. However, whereas  previous studies have assumed that language use is fixed, we address the problem of how to accommodate new language use through continual learning. 

We have proposed to learn a transformation function that adjusts the language embedding, when needed, to be closer to the image embedding of the intended referent. 
To the best of our knowledge, this problem of continual adjustment of language-image embeddings to learn new language grounding has not been addressed before. Thus, we do not have any results from prior work to compare our performance with. However, we hope that this work can serve as a benchmark for future studies and alternative models.

Returning to the four characteristics we aimed to achieve, we think that the model has shown to be \textbf{sample efficient}, as it seems to reach fairly high performance with only one training example per new category.
Second, Experiment II showed that the model was able to \textbf{generalize} from the newly learned language use, thanks to the semantic compositionality of the referring expressions, but we also saw indications that it could understand synonymous expressions to an extent. Of course, these experiments were limited to shorter expressions and more basic forms of semantic compositionality; future work should investigate to what extent the model can handle more complex constructions. 
Third, Experiment I showed that the model was fairly \textbf{robust}, as despite a slight drop in the model's original zero-shot performance, it did not exhibit catastrophic forgetting. 
Finally, the model is fairly \textbf{computationally efficient}; the transformation function uses very simple models with few parameters and the stored training examples have a very small footprint. It can thus be updated quickly on a fairly basic CPU\footnote{On an Intel Core i7-1065G7 CPU, one iteration of Experiment II (i.e., 30 model updates) takes about 1 second for the standard CoLLIE model.}. Nevertheless, since the transformation model needs to be retrained when new examples are added, there are limits to its scalability. Whether this is a problem, however, depends entirely on the use case scenario. Regardless, the continual learning of the transformation function without storing examples is also an interesting topic for future work. 

As we have seen, the scaling function plays a very important role in retaining the model's zero-shot performance, making sure that only the newly learned terms are adjusted. However, given how the scaling function was trained here (simply using common nouns as negative examples), this will not always work, and we therefore still saw a slight drop in zero-shot performance, especially as the number of new concepts to be learned increases. The scaling function could of course be more or less restrictive. For example, it could require an exact match with a training example to set the scale to 1, and 0 otherwise. This would retain all of the zero-shot performance, at the expense of being able to generalize the learning to similar language use.  Exploring more sophisticated scaling functions that provide a good balance between retention and generalization is an interesting topic for future work. For example, \shortciteA{HanulShin2017} explore the use of a generative model to generate samples for rehearsal, which alleviates the need for storing training examples. 

Of course, CoLLIE's performance also relies on the foundation model (CLIP in our case) already having good representations, and thus it is limited by the performance of this  model to accurately represent the landscape of visual properties of objects.  As pointed out by \shortciteA{Radford2021}, CLIP's representations are limited in certain aspects, including counting objects in an image or representing detailed attributes. 

In its current form, CoLLIE learns a transformation on top of the language embedding. It should in principle be possible to instead apply this transformation to the image embedding, to better match the language embedding. This choice is dependent on the use case: If it is applied to the language embedding, both the newly learned and already existing referring expressions will be correctly resolved (as we saw in Experiment I), but this would not be the case if the image embedding was transformed to match the new language use. Recent work on tuning models for image and text embeddings has also shown that it is better to freeze the image model, while fine-tuning the text model \shortcite{zhai_lit_2022}. 

An interesting topic for future work is how to consolidate the learned transformation function into the foundation model, and then learn a new transformation function on top of this, or to use different transformation functions in different contexts, as language use is highly context-dependent. Another line of future work is to incorporate the model into a system that learns through interaction. Given the small number of examples needed to learn new language use, the model should be interesting for studies on continual language grounding in the context of human-robot interaction \shortcite<e.g.,>{Chai2016}, where the robot should be able to both comprehend and generate referring expressions based on partner-specific language use.

\section{Conclusion}

We have presented CoLLIE: a simple, yet effective model for continual learning of how language is grounded in vision. Given a pre-trained language-image embedding model capable of zero-shot image classification or referring expression comprehension, such as CLIP, CoLLIE learns a transformation function that adjusts the language embeddings when needed to accommodate new language use. The transformation function learns the difference vector that needs to be applied to the embedding, and uses a scaling function to retain embeddings that should not be adjusted. We establish new benchmarks to capture the trade-off between continual learning, retention (avoiding catastrophic forgetting), and generalization. 
While our evaluation is limited in several regards, it indicates that the model can learn new language use with very few examples. Unlike traditional few-shot learning, the model does not just learn new labels, but can also generalize to similar language use, and benefit from the semantic compositionality of language. 

\section*{Reproducability Statement}

The models were implemented using scikit-learn (\url{https://scikit-learn.org/}) with standard parameters unless stated otherwise. The code used for running the experiments and reproducing the results in this paper is provided on GitHub (\url{https://github.com/gabriel-skantze/CoLLIE}), including necessary data or pointers to data.

\section*{Acknowledgements}

This work was partially supported by the Wallenberg AI, Autonomous Systems and Software Program (WASP) funded by the Knut and Alice Wallenberg Foundation. The authors would like to thank Johan Boye, Ulme Wennberg, Dmytro Kalpakchi, and the anonymous reviewers for their helpful comments.

\vskip 0.2in
\bibliography{sample}

\begin{thebibliography}{}

\bibitem[\protect\BCAY{Barr\ \BBA\ Keysar}{Barr\ \BBA\
  Keysar}{2002}]{barr_anchoring_2002}
Barr, D.~J.\BBACOMMA\  \BBA\ Keysar, B. \BBOP2002\BBCP.
\newblock \BBOQ Anchoring {Comprehension} in {Linguistic} {Precedents}\BBCQ\
\newblock {\Bem Journal of Memory and Language}, {\Bem 46\/}(2), 391--418.

\bibitem[\protect\BCAY{Benavides-Prado, Koh,\ \BBA\ Riddle}{Benavides-Prado
  et~al.}{2020}]{Benavides2020}
Benavides-Prado, D., Koh, Y.~S., \BBA\ Riddle, P. \BBOP2020\BBCP.
\newblock \BBOQ {Towards Knowledgeable Supervised Lifelong Learning
  Systems}\BBCQ\
\newblock {\Bem Journal of Artificial Intelligence Research}, {\Bem 68},
  159--224.

\bibitem[\protect\BCAY{Bommasani et~al.}{Bommasani
  et~al.}{2021}]{Bommasani2021}
Bommasani, R.\BBACOMMA\  et~al. \BBOP2021\BBCP.
\newblock \BBOQ {On the Opportunities and Risks of Foundation Models}\BBCQ\
\newblock {\Bem arXiv}, {\Bem 2108.07258}.

\bibitem[\protect\BCAY{Brennan\ \BBA\ Clark}{Brennan\ \BBA\
  Clark}{1996}]{Brennan1996}
Brennan, S.\BBACOMMA\  \BBA\ Clark, H. \BBOP1996\BBCP.
\newblock \BBOQ {Conceptual pacts and lexical choice in conversation}\BBCQ\
\newblock {\Bem Journal of Experimental Psychology}, {\Bem 22\/}(6),
  1482--1493.

\bibitem[\protect\BCAY{Brennan\ \BBA\ Hanna}{Brennan\ \BBA\
  Hanna}{2009}]{Brennan2009}
Brennan, S.~E.\BBACOMMA\  \BBA\ Hanna, J.~E. \BBOP2009\BBCP.
\newblock \BBOQ Partner-specific adaptation in dialog\BBCQ\
\newblock {\Bem Topics in Cognitive Science}, {\Bem 1\/}(2), 274--291.

\bibitem[\protect\BCAY{Bruni, Tran,\ \BBA\ Baroni}{Bruni
  et~al.}{2014}]{Bruni2014}
Bruni, E., Tran, N.~K., \BBA\ Baroni, M. \BBOP2014\BBCP.
\newblock \BBOQ {Multimodal Distributional Semantics}\BBCQ\
\newblock {\Bem Journal of Artificial Intelligence Research}, {\Bem 49}, 1--47.

\bibitem[\protect\BCAY{Chai, Fang, Liu,\ \BBA\ She}{Chai
  et~al.}{2016}]{Chai2016}
Chai, J.~Y., Fang, R., Liu, C., \BBA\ She, L. \BBOP2016\BBCP.
\newblock \BBOQ {Collaborative language grounding toward situated human-robot
  dialogue}\BBCQ\
\newblock {\Bem AI Magazine}, {\Bem 37\/}(4), 32--45.

\bibitem[\protect\BCAY{Das, Kottur, Gupta, Singh, Yadav, Moura, Parikh,\ \BBA\
  Batra}{Das et~al.}{2017}]{Das2017}
Das, A., Kottur, S., Gupta, K., Singh, A., Yadav, D., Moura, J.~M., Parikh, D.,
  \BBA\ Batra, D. \BBOP2017\BBCP.
\newblock \BBOQ {Visual dialog}\BBCQ\
\newblock In {\Bem Proceedings - 30th IEEE Conference on Computer Vision and
  Pattern Recognition, CVPR 2017}, \BPGS\ 1080--1089.

\bibitem[\protect\BCAY{Dauphin, Fan, Auli,\ \BBA\ Grangier}{Dauphin
  et~al.}{2017}]{Dauphin2017}
Dauphin, Y.~N., Fan, A., Auli, M., \BBA\ Grangier, D. \BBOP2017\BBCP.
\newblock \BBOQ Language modeling with gated convolutional networks\BBCQ\
\newblock In {\Bem Proceedings of the 34th International Conference on Machine
  Learning - Volume 70}, \BPG\ 933–941.

\bibitem[\protect\BCAY{Deng, Dong, Socher, Li, Li,\ \BBA\ Fei-Fei}{Deng
  et~al.}{2009}]{Deng2009}
Deng, J., Dong, W., Socher, R., Li, L.-J., Li, K., \BBA\ Fei-Fei, L.
  \BBOP2009\BBCP.
\newblock \BBOQ {ImageNet: A large-scale hierarchical image database}\BBCQ\
\newblock In {\Bem 2009 IEEE Conference on Computer Vision and Pattern
  Recognition}, \BPGS\ 248--255.

\bibitem[\protect\BCAY{Frome, Corrado, Shlens, Bengio, Dean, Ranzato,\ \BBA\
  Mikolov}{Frome et~al.}{2013}]{Frome2013}
Frome, A., Corrado, G.~S., Shlens, J., Bengio, S., Dean, J., Ranzato, M., \BBA\
  Mikolov, T. \BBOP2013\BBCP.
\newblock \BBOQ {DeViSE: A Deep Visual-Semantic Embedding Model}\BBCQ\
\newblock In {\Bem Proceedings of the 26th International Conference on Neural
  Information Processing Systems - Volume 2}, NIPS'13, \BPGS\ 2121--2129.

\bibitem[\protect\BCAY{Grice}{Grice}{1975}]{Grice1975LogicConversation}
Grice, H.~P. \BBOP1975\BBCP.
\newblock \BBOQ {Logic and Conversation}\BBCQ\
\newblock {\Bem Syntax and Semantics}, {\Bem 3}, 41--58.

\bibitem[\protect\BCAY{Harnad}{Harnad}{1990}]{harnad1990}
Harnad, S. \BBOP1990\BBCP.
\newblock \BBOQ The symbol grounding problem\BBCQ\
\newblock {\Bem Physica D: Nonlinear Phenomena}, {\Bem 42\/}(1-3), 335--346.

\bibitem[\protect\BCAY{He, Zhang, Ren,\ \BBA\ Sun}{He
  et~al.}{2016}]{he2016deep}
He, K., Zhang, X., Ren, S., \BBA\ Sun, J. \BBOP2016\BBCP.
\newblock \BBOQ {Deep Residual Learning for Image Recognition}\BBCQ\
\newblock In {\Bem 2016 IEEE Conference on Computer Vision and Pattern
  Recognition (CVPR)}, \BPGS\ 770--778.

\bibitem[\protect\BCAY{Horst\ \BBA\ Hout}{Horst\ \BBA\ Hout}{2016}]{Horst2016}
Horst, J.~S.\BBACOMMA\  \BBA\ Hout, M.~C. \BBOP2016\BBCP.
\newblock \BBOQ {The Novel Object and Unusual Name (NOUN) Database: A
  collection of novel images for use in experimental research}\BBCQ\
\newblock {\Bem Behavior Research Methods}, {\Bem 48\/}(4), 1393--1409.

\bibitem[\protect\BCAY{Ibarra\ \BBA\ Tanenhaus}{Ibarra\ \BBA\
  Tanenhaus}{2016}]{ibarra_flexibility_2016}
Ibarra, A.\BBACOMMA\  \BBA\ Tanenhaus, M.~K. \BBOP2016\BBCP.
\newblock \BBOQ {The Flexibility of Conceptual Pacts: Referring Expressions
  Dynamically Shift to Accommodate New Conceptualizations}\BBCQ\
\newblock {\Bem Frontiers in Psychology}, {\Bem 7}.

\bibitem[\protect\BCAY{Kafle\ \BBA\ Kanan}{Kafle\ \BBA\
  Kanan}{2017}]{Kafle2017}
Kafle, K.\BBACOMMA\  \BBA\ Kanan, C. \BBOP2017\BBCP.
\newblock \BBOQ {Visual question answering: Datasets, algorithms, and future
  challenges}\BBCQ\
\newblock {\Bem Computer Vision and Image Understanding}, {\Bem 163}, 3--20.

\bibitem[\protect\BCAY{Kemker\ \BBA\ Kanan}{Kemker\ \BBA\
  Kanan}{2018}]{kemker2018fearnet}
Kemker, R.\BBACOMMA\  \BBA\ Kanan, C. \BBOP2018\BBCP.
\newblock \BBOQ {FearNet: Brain-Inspired Model for Incremental Learning}\BBCQ\
\newblock In {\Bem Proceedings of the 6th International Conference on Learning
  Representations (ICLR 2018)}.

\bibitem[\protect\BCAY{Kemker, McClure, Abitino, Hayes,\ \BBA\ Kanan}{Kemker
  et~al.}{2018}]{kemker2018measuring}
Kemker, R., McClure, M., Abitino, A., Hayes, T., \BBA\ Kanan, C.
  \BBOP2018\BBCP.
\newblock \BBOQ {Measuring Catastrophic Forgetting in Neural Networks}\BBCQ\
\newblock In {\Bem Proceedings of the AAAI Conference on Artificial
  Intelligence}, \lowercase{\BVOL}~32.

\bibitem[\protect\BCAY{Krahmer\ \BBA\ van Deemter}{Krahmer\ \BBA\ van
  Deemter}{2012}]{Krahmer2012}
Krahmer, E.\BBACOMMA\  \BBA\ van Deemter, K. \BBOP2012\BBCP.
\newblock \BBOQ {Computational Generation of Referring Expressions: A
  Survey}\BBCQ\
\newblock {\Bem Computational Linguistics}, {\Bem 38\/}(1), 173--218.

\bibitem[\protect\BCAY{McCloskey\ \BBA\ Cohen}{McCloskey\ \BBA\
  Cohen}{1989}]{mccloskey1989}
McCloskey, M.\BBACOMMA\  \BBA\ Cohen, N.~J. \BBOP1989\BBCP.
\newblock \BBOQ {Catastrophic Interference in Connectionist Networks: The
  Sequential Learning Problem}\BBCQ\
\newblock {\Bem Psychology of Learning and Motivation}, {\Bem 24}, 109--165.

\bibitem[\protect\BCAY{Panagiaris, Hart,\ \BBA\ Gkatzia}{Panagiaris
  et~al.}{2021}]{Panagiaris2021}
Panagiaris, N., Hart, E., \BBA\ Gkatzia, D. \BBOP2021\BBCP.
\newblock \BBOQ {Generating unambiguous and diverse referring
  expressions}\BBCQ\
\newblock {\Bem Computer Speech and Language}, {\Bem 68}, 101184.

\bibitem[\protect\BCAY{Parisi, Kemker, Part, Kanan,\ \BBA\ Wermter}{Parisi
  et~al.}{2019}]{Parisi2019}
Parisi, G.~I., Kemker, R., Part, J.~L., Kanan, C., \BBA\ Wermter, S.
  \BBOP2019\BBCP.
\newblock \BBOQ {Continual lifelong learning with neural networks: A
  review}\BBCQ\
\newblock {\Bem Neural Networks}, {\Bem 113}, 54--71.

\bibitem[\protect\BCAY{Pedregosa, Varoquaux, Gramfort, Michel, Thirion, Grisel,
  Blondel, Prettenhofer, Weiss, Dubourg, Vanderplas, Passos, Cournapeau,
  Brucher, Perrot,\ \BBA\ Duchesnay}{Pedregosa et~al.}{2011}]{scikit-learn}
Pedregosa, F., Varoquaux, G., Gramfort, A., Michel, V., Thirion, B., Grisel,
  O., Blondel, M., Prettenhofer, P., Weiss, R., Dubourg, V., Vanderplas, J.,
  Passos, A., Cournapeau, D., Brucher, M., Perrot, M., \BBA\ Duchesnay, E.
  \BBOP2011\BBCP.
\newblock \BBOQ Scikit-learn: Machine {L}earning in {P}ython\BBCQ\
\newblock {\Bem Journal of Machine Learning Research}, {\Bem 12}, 2825--2830.

\bibitem[\protect\BCAY{Pickering\ \BBA\ Garrod}{Pickering\ \BBA\
  Garrod}{2006}]{Pickering2006}
Pickering, M.\BBACOMMA\  \BBA\ Garrod, S. \BBOP2006\BBCP.
\newblock \BBOQ {Alignment as the Basis for Successful Communication}\BBCQ\
\newblock {\Bem Research on Language and Computation}, {\Bem 4}, 203--228.

\bibitem[\protect\BCAY{Qiao, Deng,\ \BBA\ Wu}{Qiao et~al.}{2021}]{Qiao2020}
Qiao, Y., Deng, C., \BBA\ Wu, Q. \BBOP2021\BBCP.
\newblock \BBOQ {Referring Expression Comprehension: A Survey of Methods and
  Datasets}\BBCQ\
\newblock {\Bem IEEE Transactions on Multimedia}, {\Bem 23}, 4426--4440.

\bibitem[\protect\BCAY{Radford, Kim, Hallacy, Ramesh, Goh, Agarwal, Sastry,
  Askell, Mishkin, Clark, Krueger,\ \BBA\ Sutskever}{Radford
  et~al.}{2021}]{Radford2021}
Radford, A., Kim, J.~W., Hallacy, C., Ramesh, A., Goh, G., Agarwal, S., Sastry,
  G., Askell, A., Mishkin, P., Clark, J., Krueger, G., \BBA\ Sutskever, I.
  \BBOP2021\BBCP.
\newblock \BBOQ {Learning Transferable Visual Models From Natural Language
  Supervision}\BBCQ\
\newblock {\Bem arXiv}, {\Bem 2103.00020}.

\bibitem[\protect\BCAY{Rebuffi, Kolesnikov, Sperl,\ \BBA\ Lampert}{Rebuffi
  et~al.}{2017}]{rebuffi2017icarl}
Rebuffi, S.-A., Kolesnikov, A., Sperl, G., \BBA\ Lampert, C.~H. \BBOP2017\BBCP.
\newblock \BBOQ {iCaRL: Incremental Classifier and Representation
  Learning}\BBCQ\
\newblock In {\Bem Proceedings of the IEEE Conference on Computer Vision and
  Pattern Recognition (CVPR)}, \BPGS\ 2001--2010.

\bibitem[\protect\BCAY{Shin, Lee, Kim,\ \BBA\ Kim}{Shin
  et~al.}{2017}]{HanulShin2017}
Shin, H., Lee, J.~K., Kim, J., \BBA\ Kim, J. \BBOP2017\BBCP.
\newblock \BBOQ {Continual Learning with Deep Generative Replay}\BBCQ\
\newblock In {\Bem NIPS'17: Proceedings of the 31st International Conference on
  Neural Information Processing Systems}, \BPGS\ 2994--3003.

\bibitem[\protect\BCAY{Shore, Androulakaki,\ \BBA\ Skantze}{Shore
  et~al.}{2018}]{Shore2018-LREC}
Shore, T., Androulakaki, T., \BBA\ Skantze, G. \BBOP2018\BBCP.
\newblock \BBOQ {KTH Tangrams: A Dataset for Research on Alignment and
  Conceptual Pacts in Task-Oriented Dialogue}\BBCQ\
\newblock In {\Bem LREC 2018 - 11th International Conference on Language
  Resources and Evaluation}, \BPGS\ 768--775.

\bibitem[\protect\BCAY{Shore\ \BBA\ Skantze}{Shore\ \BBA\
  Skantze}{2018}]{Shore2018-EMNLP}
Shore, T.\BBACOMMA\  \BBA\ Skantze, G. \BBOP2018\BBCP.
\newblock \BBOQ {Using Lexical Alignment and Referring Ability to Address Data
  Sparsity in Situated Dialog Reference Resolution}\BBCQ\
\newblock In {\Bem Proceedings of the 2018 Conference on Empirical Methods in
  Natural Language Processing (EMNLP)}, \BPGS\ 2288--2297.

\bibitem[\protect\BCAY{Takmaz, Giulianelli, Pezzelle, Sinclair,\ \BBA\
  Fern{\'{a}}ndez}{Takmaz et~al.}{2020}]{Takmaz2020}
Takmaz, E., Giulianelli, M., Pezzelle, S., Sinclair, A., \BBA\ Fern{\'{a}}ndez,
  R. \BBOP2020\BBCP.
\newblock \BBOQ {Refer, Reuse, Reduce: Generating Subsequent References in
  Visual and Conversational Contexts}\BBCQ\
\newblock In {\Bem Proceedings of the 2020 Conference on Empirical Methods in
  Natural Language Processing (EMNLP)}, \BPGS\ 4350--4368.

\bibitem[\protect\BCAY{{Van Der Maaten}\ \BBA\ Hinton}{{Van Der Maaten}\ \BBA\
  Hinton}{2008}]{VanDerMaaten2008}
{Van Der Maaten}, L.\BBACOMMA\  \BBA\ Hinton, G. \BBOP2008\BBCP.
\newblock \BBOQ {Visualizing Data using t-SNE}\BBCQ\
\newblock {\Bem Journal of Machine Learning Research}, {\Bem 9\/}(86),
  2579--2605.

\bibitem[\protect\BCAY{Wang, Yao, Kwok,\ \BBA\ Ni}{Wang
  et~al.}{2020}]{Wang2020}
Wang, Y., Yao, Q., Kwok, J.~T., \BBA\ Ni, L.~M. \BBOP2020\BBCP.
\newblock \BBOQ {Generalizing from a Few Examples: A Survey on Few-Shot
  Learning}\BBCQ\
\newblock {\Bem ACM Computing Surveys}, {\Bem 53\/}(3), 1--34.

\bibitem[\protect\BCAY{You, Jin, Wang, Fang,\ \BBA\ Luo}{You
  et~al.}{2016}]{You2016}
You, Q., Jin, H., Wang, Z., Fang, C., \BBA\ Luo, J. \BBOP2016\BBCP.
\newblock \BBOQ {Image Captioning With Semantic Attention}\BBCQ\
\newblock In {\Bem Proceedings of the IEEE Conference on Computer Vision and
  Pattern Recognition (CVPR)}, \BPGS\ 4651--4659.

\bibitem[\protect\BCAY{Zhai, Wang, Mustafa, Steiner, Keysers, Kolesnikov,\
  \BBA\ Beyer}{Zhai et~al.}{2022}]{zhai_lit_2022}
Zhai, X., Wang, X., Mustafa, B., Steiner, A., Keysers, D., Kolesnikov, A.,
  \BBA\ Beyer, L. \BBOP2022\BBCP.
\newblock \BBOQ {LiT}: {Zero}-{Shot} {Transfer} with {Locked}-image text
  {Tuning}\BBCQ\
\newblock {\Bem arXiv}, {\Bem 2111.07991}.

\bibitem[\protect\BCAY{Zhao, Fu, Liang, Wu, Wang,\ \BBA\ Wang}{Zhao
  et~al.}{2018}]{Zhao2018}
Zhao, B., Fu, Y., Liang, R., Wu, J., Wang, Y., \BBA\ Wang, Y. \BBOP2018\BBCP.
\newblock \BBOQ {A Large-scale Attribute Dataset for Zero-shot Learning}\BBCQ\
\newblock {\Bem arXiv}, {\Bem 1804.04314}.

\end{thebibliography}
\bibliographystyle{theapa}

\end{document}